\documentclass[twocolumn, 10pt]{article}
\usepackage{graphicx} 

\usepackage{cite}
\usepackage{amsmath,amssymb,amsfonts}
\usepackage{algorithmic}
\usepackage{graphicx}
\usepackage{textcomp}
\usepackage{xcolor}
\usepackage{float}
\usepackage{comment}
\usepackage{enumitem}
\usepackage{amssymb}
\usepackage{mathrsfs}
\usepackage{amsmath}
\usepackage{geometry}
\usepackage{setspace}
\usepackage{enumitem}
\usepackage{amssymb}
\usepackage{setspace}
\usepackage{adjustbox}
\usepackage{float}
\usepackage{booktabs}
\usepackage{amsmath}
\usepackage{xcolor}
\usepackage{acronym}
\usepackage{url}
\usepackage{relsize}
\usepackage{caption}
\usepackage{subcaption}
\usepackage{tabularx}

\makeatletter
\renewenvironment{abstract}{%
  \par\small
  \noindent\textbf{Abstract—}\ignorespaces
}{%
  \par\normalsize\bfseries
}
\makeatother

\usepackage{tikz}

\newcommand{\copyrightnotice}{
\begin{tikzpicture}[remember picture,overlay]
\node[anchor=south,yshift=10pt] at (current page.south) {
\fbox{\parbox{\dimexpr\textwidth-2\fboxsep-2\fboxrule\relax}{
\footnotesize
© 2026 IEEE. Personal use of this material is permitted. Permission from IEEE must be obtained for all other uses, in any current or future media, including reprinting/republishing this material for advertising or promotional purposes, creating new collective works, for resale or redistribution to servers or lists, or reuse of any copyrighted component of this work in other works.

This is the author’s version of a paper accepted for publication in the \href{https://doi.org/10.1109/MELECON64486.2026}{2026 IEEE 23rd Mediterranean Electrotechnical Conference (MELECON)}. The published version is available via DOI: \href{https://doi.org/10.1109/MELECON64486.2026.11418865}{10.1109/MELECON64486.2026.11418865}.
}}
};
\end{tikzpicture}
}

\usepackage{titlesec}

\renewcommand{\thesection}{\Roman{section}}
\renewcommand{\thesubsection}{\Alph{subsection}}
\renewcommand{\thesubsubsection}{\arabic{subsubsection}}

\titleformat{\section}
  {\normalfont\filcenter\MakeUppercase}
  {\thesection.}
  {0.6em}
  {}

\titleformat{\subsection}
  {\normalfont\itshape}
  {\thesubsection.}
  {0.6em}
  {}

\titleformat{\subsubsection}[runin]
  {\normalfont\itshape}
  {\thesubsubsection)}
  {0.6em}
  {}
  [.\ ] 

\titlespacing*{\section}{0pt}{1.2ex plus 0.3ex minus 0.2ex}{0.8ex}
\titlespacing*{\subsection}{0pt}{1.0ex plus 0.3ex minus 0.2ex}{0.6ex}
\titlespacing*{\subsubsection}{0pt}{0.8ex plus 0.2ex minus 0.2ex}{0.6em}

\usepackage[pdfborder={0 0 0}, colorlinks=true, linkcolor=black, citecolor=black, urlcolor= black]{hyperref}
\def\BibTeX{{\rm B\kern-.05em{\sc i\kern-.025em b}\kern-.08em
    T\kern-.1667em\lower.7ex\hbox{E}\kern-.125emX}}

\begin{document}

\title{\huge\bfseries Network Distributed Multi-Agent Reinforcement Learning for Consensus Control of Quadcopters}

\author{
\begin{tabular}{c}
\normalsize
Youssef Mahran$^{1}$, Zeyad Gamal$^{1}$, Aamir Ahmad$^{2}$, and Ayman A. El-Badawy$^{3}$\\[0.6em]
\normalsize
$^{1}$Mechatronics Engineering Department, German University in Cairo (GUC), Egypt\\
\normalsize $^{2}$Institute of Flight Mechanics and Control (IFR), Head of Flight Robotics, University of Stuttgart, Germany\\
\normalsize $^{3}$Faculty of EMS, Head of Mechatronics Engineering Department, German University in Cairo (GUC), Egypt\\[0.6em]
\normalsize
Emails: youssef.mahran@student.guc.edu.eg, zeyad.gamal@guc.edu.eg,\\ 
\normalsize aamir.ahmad@ifr.uni-stuttgart.de, ayman.elbadawy@guc.edu.eg
\end{tabular}
\thanks{ $^2$This work has been supported by the "Optimal Networked Control" Project funded by the German Academic Exchange Service (DAAD) and the Federal Ministry for Education, Research, Technology and Aerospace (BMFTR) in the framework of the DAAD-TNB-BINA funding project "GUC: Building A Sustainable Future" (project ID: 57710434).}}

\date{} 

\maketitle

\copyrightnotice
\begin{abstract}
\bfseries This paper proposes a Network Distributed Multi-Agent Reinforcement Learning (ND-MARL) framework for quadcopter consensus control. Compared to conventional multi-agent MARL formulations that rely on centralized planning or fully decentralized execution, ND-MARL incorporates the swarm communication graph into the decision process. Under a 2-Neighbor communication topology, each agent observes information of only two neighbors and outputs an action through a distributed policy. A high-level distributed consensus planner is trained using Multi-Agent Soft Actor-Critic (MASAC) and embedded in a hierarchical stack to generate reference target positions tracked by a low-level quadcopter controller. Results demonstrate smooth consensus trajectories and planner-tracker integration when compared to a centralized MARL controller. Most notably, the learned controller exhibits zero-shot scalability, as policies trained on a three-agent system are deployed to swarms of up to 250 agents under the same 2-Neighbor communication topology without retraining or fine-tuning, achieving consistent convergence with increasing steady-state spread at large team sizes due to sparse information propagation. These findings highlight ND-MARL as a stable framework for distributed, communication-aware quadcopter consensus control.
\end{abstract}

\section{Introduction}
Multi-UAV systems are increasingly central to aerial robotics, enabling missions such as disaster response, surveillance, mapping, and target tracking to be carried out more efficiently than by single platforms. By operating in coordinated teams, quadcopters can extend coverage areas, enhance robustness through redundancy, and achieve scalability to larger and more complex mission spaces. At the same time, practical deployment of UAV swarms requires solving issues such as real-time collision avoidance, decentralized decision-making, and consensus on shared objectives \cite{Nouran,loaie,Ali,mokhtar2023autonomous}.

Multi-agent reinforcement learning (MARL) has emerged as a powerful framework for coordinating UAV swarms, with existing approaches broadly categorized into centralized, decentralized, and distributed paradigms. Centralized MARL formulations treat the entire swarm as a single system, where a joint controller observes the global state and outputs coordinated actions for all UAVs. While this guarantees consistency and avoids non-stationarity during learning, centralized approaches scale poorly due to the exponential growth of the joint state-action space and the heavy communication required to maintain global information \cite{Bernstein2002,Ekechi2025}. These limitations restrict centralized MARL to small teams or simulated environments, making it unsuitable for large-scale UAV swarms.

Decentralized MARL removes reliance on a central controller by allowing each UAV to learn its own policy from local observations. This improves scalability and robustness at execution time but introduces significant learning challenges. In particular, simultaneous policy updates by multiple agents create a non-stationary environment from each agent’s perspective, often leading to unstable or inefficient learning \cite{HernandezLeal2019}. To address this issue, the centralized training with decentralized execution (CTDE) paradigm was introduced, enabling agents to leverage global state or joint action information during training while remaining fully decentralized during execution \cite{lowe2017multi,Foerster2018}. CTDE methods have demonstrated strong performance in cooperative UAV tasks; however, their reliance on centralized critics during training limits scalability as swarm size increases \cite{HernandezLeal2019,Ekechi2025}. 
\begin{figure*}[!t]
    \centering
    \includegraphics[width=0.95\textwidth]{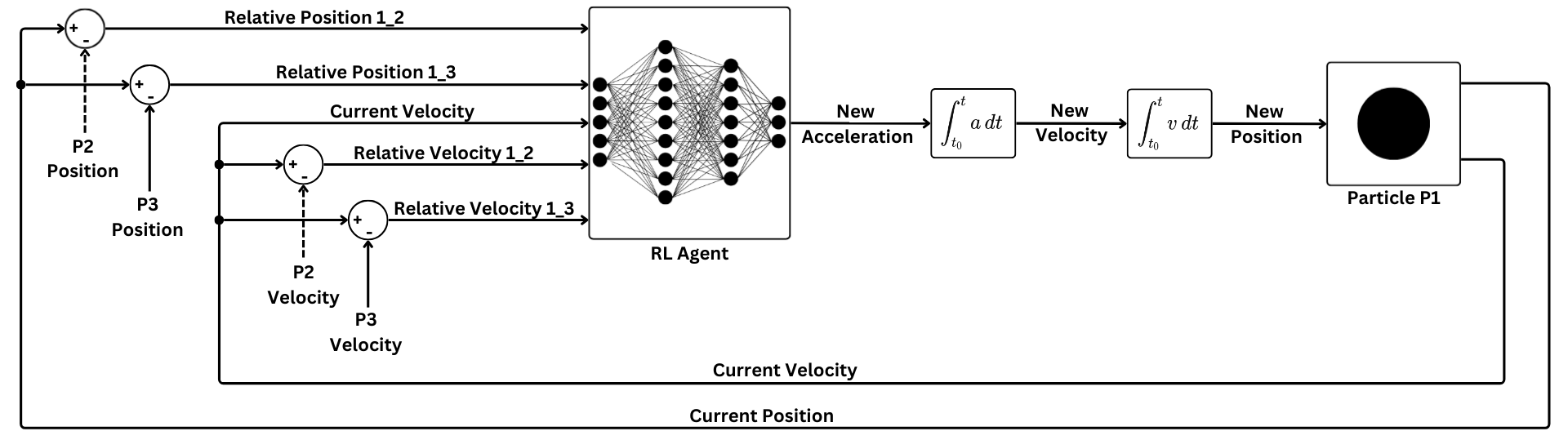}
    \caption{Single-particle consensus planner for one agent. The policy maps neighbor-relative positions and velocities to planar accelerations, which are integrated twice to update the particle position. This abstraction isolates the high-level consensus policy learned in the continuous ND-MARL before embedding it into the full quadcopter stack in Fig.~\ref{fig:rl_stack}.}
    \label{fig:cons_particle}
\end{figure*}

Distributed MARL provides an intermediate solution by enabling UAVs to act based on local observations while explicitly communicating with neighboring agents. By structuring interactions through local communication graphs, distributed approaches reduce dependence on global information and better reflect real-world communication constraints. Early work demonstrated that learnable communication protocols significantly enhance cooperative behavior \cite{Sukhbaatar2016,Foerster2016}, with later studies introducing attention-based and targeted communication mechanisms to improve efficiency \cite{Jiang2018,Das2019}. In UAV swarm applications, graph-based and nearest-neighbor communication strategies have shown improved scalability and robustness compared to centralized and CTDE methods \cite{Khan2020}. Mean-field approximations partially address scalability by reducing interaction complexity but may oversimplify coordination in tightly coupled UAV tasks \cite{Yang2018}. These limitations highlight the need for distributed MARL frameworks that achieve scalable coordination through localized interactions while maintaining stability and generalization.

The aim of this paper is to develop a Network Distributed Multi-Agent Reinforcement Learning (ND-MARL) framework for UAV consensus, where each quadcopter communicates only with its two neighbors. By structuring the problem as a set of local pairwise agreements rather than requiring full global information, the framework addresses the scalability limits of centralized approaches while avoiding the instability of purely decentralized methods and achieves zero-shot scalability without further training.

\section{Methodology}
\subsection{Multi-Agent Quadcopter Consensus Planner}
\label{chp:marl}

The consensus problem is modeled as an ND-MARL problem, where a set of agents $\mathcal{I}=\{1,\dots,N\}$ have an undirected communication graph $\mathcal{G}=(\mathcal{I},\mathcal{E})$ with edge set $\mathcal{E}$. This breaks the global consensus problem into smaller subproblems within each neighborhood. 

Agents are modeled as particles with simple kinematics. Each agent executes a local policy using only information from the fixed-size neighborhood induced by $\mathcal{G}$ as shown in Fig.~\ref{fig:comms}. 

The learned particle-level consensus policy is then embedded as a high-level planner in a hierarchical quadcopter control stack, where it outputs reference targets that are tracked by low-level single-agent RL quadcopter controllers.

\subsection{2-Neighbor Graph Communication Topology ($\mathcal{G}$)}
\begin{figure}[H]
    \centering
    \includegraphics[width=0.35\textwidth]{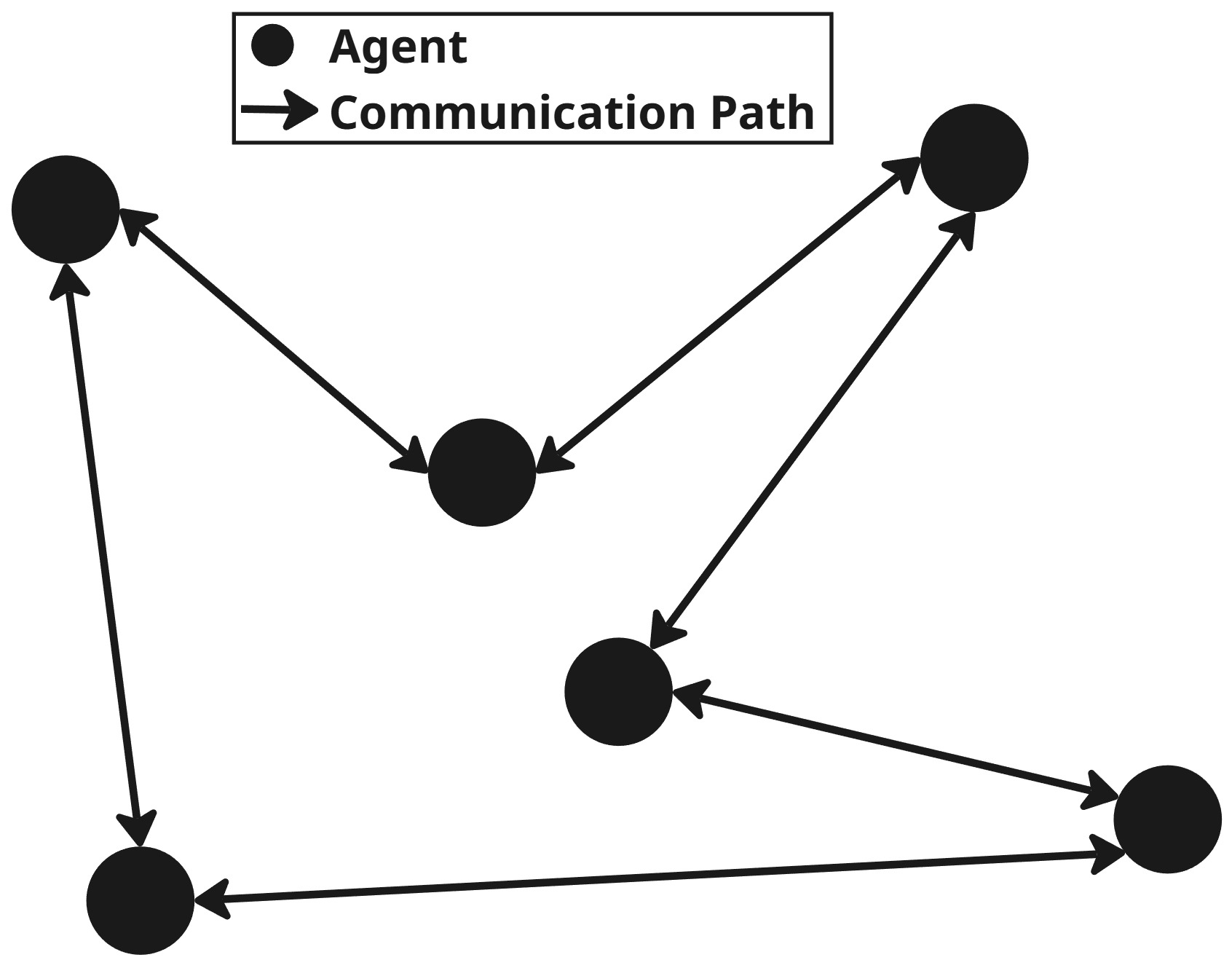}
    \caption{A set of 6 agents under an undirected 2-Neighbor symmetric communication topology, with each agent exchanging information with only its two neighbors}
    \label{fig:comms}
\end{figure}

A critical design choice in the proposed framework is
the communication topology among agents. Instead of assuming
full observability, a fixed two-degree communication topology is applied in which each agent $i$ maintains exactly two symmetric links as shown in Fig.~\ref{fig:comms}. Let $\mathcal{N}(i)$ denote the neighbor set of agent $i$ then $|\mathcal{N}(i)|=2$ and $(i,j)\in\mathcal{E}$ implies $(j,i)\in\mathcal{E}$. This constant-degree design preserves the locality structure and enables distributed execution with per-agent computation and communication that do not grow with $N$.

Under the locality assumption, the global reward ($\mathcal{R}_G$) factorizes with respect to agent neighborhoods as shown in Eq.~\eqref{eq:rew_fact}.
\begin{equation}
\mathcal{R}_G(s,a) = \sum_{i\in\mathcal{I}}
\mathcal{R}_{i}\!\left(s_i, s_{\mathcal{N}(i)}, a_i, a_{\mathcal{N}(i)}\right)
\label{eq:rew_fact}
\end{equation}

In the ND-MARL implementation proposed, each agent observes only relative states (e.g., relative positions and velocities) with respect to its two neighbors and outputs a local action. This localized topology provides: 
\begin{enumerate}
    \item Scalability, as the communication and computational cost per agent remains constant regardless of the swarm size
    \item Distributed coordination, because policies leverage neighbor information to achieve consensus globally.
\end{enumerate}

\subsection{Consensus Path Planner Action ($\mathcal{A}$) and State ($\mathcal{S}$) Spaces}
\begin{figure}[H]
    \centering
    \includegraphics[width=0.4\textwidth]{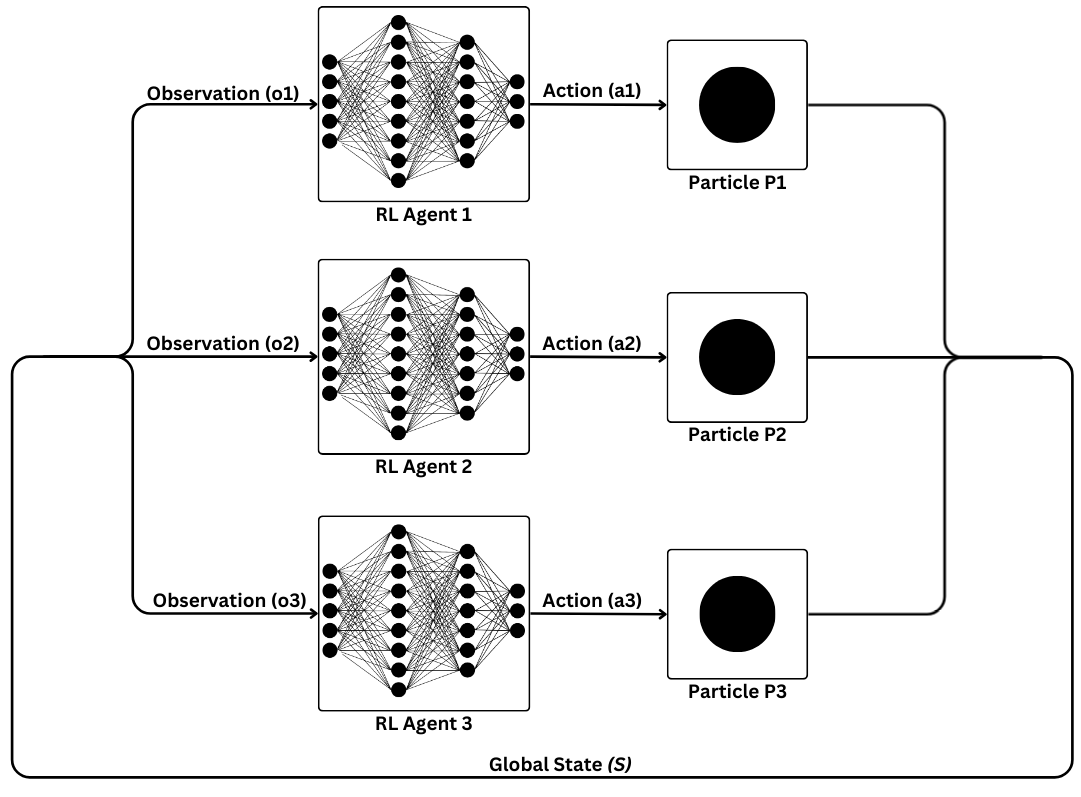}
    \caption{Multi-agent consensus path-planner architecture. Each agent runs a separate MASAC policy using only relative information from its two neighbors to drive the cluster to consensus over the 2-Neighbor communication graph.}
    \label{fig:cons_cluster}
\end{figure}

Each agent controls a massless particle via a high-level MARL policy, as shown in Fig.~\ref{fig:cons_cluster}. At each step, agent $i$ outputs a planar acceleration for its particle, forming the action space $\mathcal{A}$ in Eq.~\eqref{eq:action_space}, which is integrated twice to update the particle position as shown in Fig.~\ref{fig:cons_particle}.
\begin{subequations}
    \begin{align}
    \mathcal{A}_i &= [a_x, a_y]\\
    a_x,a_y &\in [-1,1]
    \end{align}
    \label{eq:action_space}
\end{subequations}
Agents observe relative positions ($p_{ij}$) and velocities ($v_{ij}$) of neighbors and their own velocity ($v_x, v_y$), forming their state space $\mathcal{S}_i$ as shown in Eq.~\eqref{eq:obs_space} with all quantities expressed relative to agent $i$. This makes the environment translation-invariant, as the agents are unaware of their absolute positions and can form consensus regardless of the cluster position.

\begin{subequations}
\label{eq:obs_space}
\begin{align}
    \mathcal{S}_i &= [v_x, v_y, p_{ij}, p_{ik}, v_{ij}, v_{ik}] \label{eq:obs_space_main} \\
    p_{ij} &= [x_{ij}, y_{ij}] \label{eq:obs_space_p} \\
    v_{ij} &= [v_{x_{ij}}, v_{y_{ij}}] \label{eq:obs_space_v}\\
    (j,k) &\in \mathcal{N}(i)\\
    \mathcal{S}_i &\in [-1,1]
\end{align}
\end{subequations}
\subsection{Reward Function ($\mathcal{R}_i$)}
\begin{figure*}[!t]
    \centering
    \includegraphics[width=0.925\linewidth]{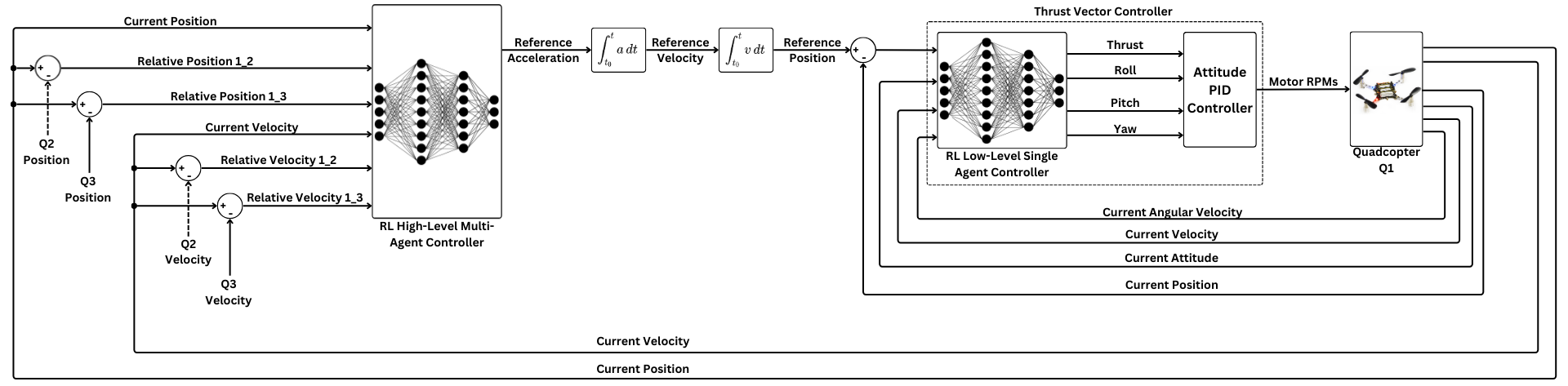}
    \caption{Hierarchical reinforcement learning stack for a single quadcopter. A high-level multi-agent consensus planner outputs target positions based on neighbor-relative information, while a low-level single-agent RL thrust-vector controller \cite{niles} converts position errors into thrust and attitude commands. An attitude PID loop tracks these commands, yielding hardware-realistic quadcopter motion that inherits the learned consensus behavior.}
    \label{fig:rl_stack}
\end{figure*}
\label{sec:rew}
Each agent receives a local reward combining dense consensus and velocity terms with a sparse consensus-loss penalty. The mean distance of agent $i$ relative to neighbors normalized by the workspace diagonal is shown in Eq.~\eqref{eq:mean_dist}. Where $|\mathcal{N}(i)|$ is the number of neighbors of agent $i$, $p$ is the absolute position and $B_e = 3 m$ is the environment bounds.
\begin{subequations}
    \begin{align}
    \bar{d}_i = \frac{1}{|\mathcal{N}(i)|} &\sum_{j \in \mathcal{N}(i)} \lVert p_i - p_j \rVert\\
    d_{norm} &= 2B_e \sqrt{2} \\
    \tilde{d}_i &= \frac{\bar{d}_i}{d_{norm}}
    \end{align}
    \label{eq:mean_dist}
\end{subequations}

The consensus reward is bounded to the range [0,1) by the $\tanh$ function as shown in Eq.~\eqref{eq:r_cons} where $\kappa_c = 0.15$ is a scaling parameter controlling sensitivity.
\begin{equation}
    r^i_{cons} = 1 - \tanh\!\left(\frac{\tilde{d}_i}{\kappa_c}\right)
    \label{eq:r_cons}
\end{equation}

Consensus is declared if $\bar{d}_{nh}$, shown in Eq.~\eqref{eq:global_mean}, falls below a threshold $\delta_c = 0.2 \ m$. Note that $\bar{d}_{nh}$ measures the pairwise spread among the neighbors themselves, whereas the earlier quantity $\bar{d}_i$ in Eq.~\eqref{eq:mean_dist} measures the average pairwise distance from agent $i$ to its neighbors. Thus, $\bar{d}_i$ captures how close $i$ is to its neighborhood $\mathcal{N}(i)$, while $\bar{d}_{nh}$ captures how tightly that neighborhood is clustered.

\begin{equation}
\bar d_{{nh}_i}
=\frac{2}{\big(|\mathcal{N}(i)|+1\big)\,|\mathcal{N}(i)|}
\sum_{\substack{j<k\\ j,k\in \mathcal{N}(i)\cup\{i\}}}\left\lVert p_j-p_k \right\rVert
\label{eq:global_mean}
\end{equation}

A conditional velocity reward shown in Eq.~\eqref{eq:r_vel} encourages stationarity after consensus, where $v_i$ is the absolute velocity of agent $i$, $v_{max}$ is the maximum velocity and $\kappa_v = 0.2$ is a scaling parameter. This reward is only activated after consensus is declared. This term is introduced to eliminate a residual drift i the particles that remained after consensus.

\begin{equation}
r^i_{vel} = 1 - \tanh\left(\frac{\lVert v_i \rVert / v_{max}}{\kappa_v}\right)
\label{eq:r_vel}
\end{equation}

 After consensus, loss of consensus induces a sparse penalty shown in Eq.~\eqref{eq:r_pen}.
 
\begin{equation}
    r_{pen} =
    \begin{cases} 
        -1, & \text{if consensus was achieved and} \, \bar{d}_{nh}> \delta_c \\
         0, & \text{otherwise}
    \end{cases}
    \label{eq:r_pen}
\end{equation}

The total local reward is shown in Eq.~\eqref{eq:final_reward} with the global reward being $\mathcal{R}_G = \sum_{i \in \mathcal{I}} \mathcal{R}_i$, $w_c = 1.0$ and $w_v = 0.5$. Note that the factorization condition of Eq.~\eqref{eq:rew_fact} holds, as the local reward of agent $i$ is only affected by its state and its neighbors' states. 

\begin{equation}
    \mathcal{R}_i = w_c \, r^i_{cons} + w_v \, r^i_{vel} + r_{pen}
    \label{eq:final_reward}
\end{equation}

This structure yields an aggregation phase (distance minimization) followed by a stabilization phase (velocity damping and consensus maintenance), while remaining translationally invariant.

\subsection{Centralized Controller Baseline}
In order to evaluate the proposed scalable ND-MARL approach, a centralized MARL controller is derived to serve as a baseline. The state space $\mathcal{S}^{c}$ is the joint state for $N$ agents, and the action space $\mathcal{A}^{c}$ is the joint action, as shown in Eq.~\eqref{eq:central_state_space}.
\begin{subequations}
\label{eq:central_state_space}
\begin{align}
    \mathcal{S}^{c} &= \prod_{i \in \mathcal{I}} \mathcal{S}_i^{c} \label{eq:central_state_prod}\\
    \mathcal{S}_i^{c} &= [v_{x_i}, v_{y_i}, \{p_{ij}\}_{j \in \mathcal{I}> i}] 
    \label{eq:central_state_main}\\
    \mathcal{A}^{c} &= \prod_{i \in \mathcal{I}} \mathcal{A}_i^{c} \label{eq:central_action_prod}\\
    \mathcal{A}_i^{c} &= [a_{x_i}, a_{y_i}] \label{eq:central_action_main}
\end{align}
\end{subequations}
As the central controller has access to all velocities of all agents, the relative velocities were removed as they introduce redundancy and unnecessary complexity. 

In contrast to the distributed formulation in Sec.~\ref{sec:rew}, where each agent receives a local reward $\mathcal{R}_i$ computed only over its neighborhood $\mathcal{N}(i)$ in Eqs.~\eqref{eq:mean_dist}–\eqref{eq:final_reward}, the centralized counterpart receives the mean global reward as shown in Eq.~\eqref{eq:central_rew} where $\mathcal{R}_i$ is the reward in Eq.~\eqref{eq:final_reward}.
\begin{equation}
    \mathcal{R}_G = \frac{1}{N}\sum_{i\in\mathcal{I}} \mathcal{R}_i
    \label{eq:central_rew}
\end{equation}

\subsection{Network Architecture and Hyperparameters}

The high-level planner is trained using Multi-Agent Soft Actor-Critic (MASAC), as its entropy maximization encourages stochastic exploration, prevents premature convergence, and enhances robustness in uncertain swarm environments \cite{yue2023fmasac}, and our prior work further proved that dynamic entropy tuning improves stability in quadcopter control \cite{entropy}. We use a compact MLP with two hidden layers of 256 nodes and ReLU activations for the policy and critic networks. The main hyperparameters are summarized in Table~\ref{tab:marl_hyperparameters}.

\begin{table}[H]
    \centering
    \caption{Hyperparameters for the MASAC high-level planner}
    \begin{tabular}{lll}
    \hline
    Symbol & Description & Value \\
    \hline
    $\lambda$ & Learning rate & $0.001$ \\
    $\mathbf{R}$ & Replay buffer size &  $1{,}000{,}000$\\
    $\mathcal{B}$ & Minibatch size & $512$ \\
    $\tau$ & Target update coefficient  & $0.01$ \\
    $\gamma$ & Discount factor & $0.99$ \\
    $-$ & Training frequency  & 8 ep  \\
    $t_{max}$  & Max steps per episode  & $800$ \\
    $v_{max}$  & Maximum Particle Velocity  & $8.0 \ \mathrm{m/s}$ \\
    $a_{max}$  & Maximum Particle Acceleration  & $4.0 \ \mathrm{m/s^2}$ \\
    $dt$ & Integration Step Size & $0.05 \ \mathrm{s}$\\
    $H_{target}$ & Target entropy & $-2$ \\
    $\alpha_0$ & Initial entropy coefficient & $0.2$ \\
    \hline
    \end{tabular}
    \label{tab:marl_hyperparameters}
\end{table}

\subsection{Hierarchical RL Quadcopter Control}
The particle-level consensus planner is embedded into a hierarchical RL stack for each quadcopter, as shown in Fig.~\ref{fig:rl_stack}. The high-level MARL planner maps the observations in Eq.~\eqref{eq:obs_space} to target positions, whereas a low-level thrust-vector controller, derived in our previous work~\cite{niles}, tracks these targets. The quadcopters are simulated inside the gym-pybullet-drones \cite{panerati2021learning} python environment.

The action space of the low-level controller is shown in Eq.~\eqref{eq:low_act} where $U$ is the total thrust and $(\phi,\theta)$ are roll/pitch commands. These commands, together with the current yaw $\psi$, are fed to an attitude PID that generates motor commands.
\begin{subequations}
    \begin{align}
        A &= [U, \phi, \theta]\\
        A &\in [-1,1]
    \end{align}
    \label{eq:low_act}
\end{subequations}
The low-level controller state is shown in Eq.~\eqref{state}.  It includes the quadcopter’s orientation ($\phi, \theta, \psi$), linear velocities ($v_x, v_y, v_z$), angular velocities ($\omega_x, \omega_y, \omega_z$), and the position error relative to the target ($\Delta x, \Delta y, \Delta z$), respectively.
\begin{subequations}
    \begin{align}
        S = [\phi, \theta, \psi, v_x, v_y, &v_z, \omega_x, \omega_y, \omega_z, \Delta x, \Delta y, \Delta z]\\
        S &\in [-1,1]
    \end{align}
\label{state}
\end{subequations}
This hierarchical design separates high-level decision making from low-level actuation, enabling scalable MARL-based consensus formation while ensuring that the quadcopters remain dynamically feasible under low-level control.

\section{Results}
\subsection{High-Level Consensus Path Planner}
\begin{figure}[H]
\centering
\includegraphics[width=0.5\textwidth]{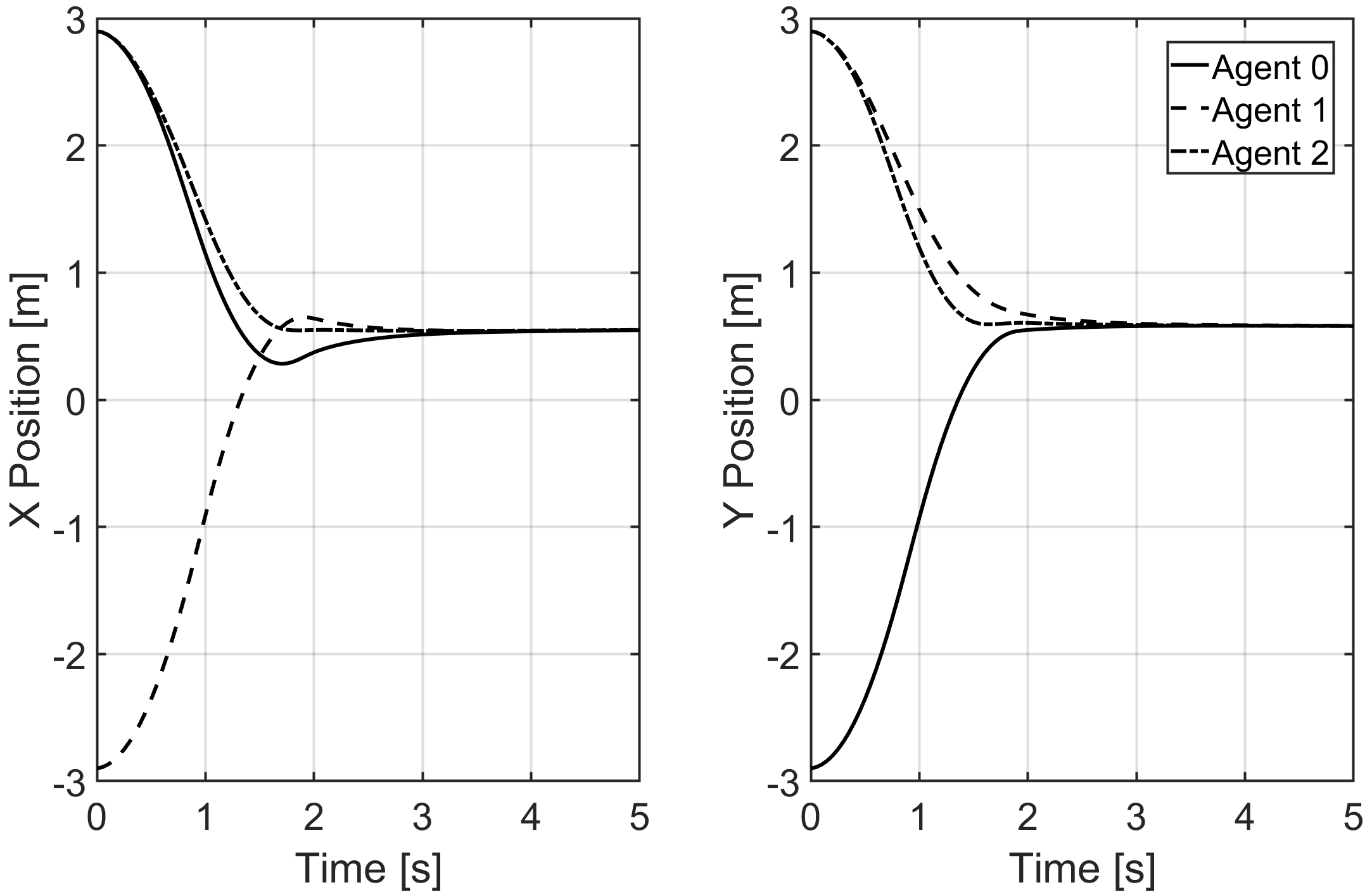}
\caption{Component-wise position responses along the $x$- and $y$-axes for three agents driven by the MASAC high-level consensus planner under a 2-Neighbor communication topology converging to a common position}
\label{fig:sac_xy}
\end{figure}

Fig.~\ref{fig:sac_xy} shows the planar trajectory produced by the proposed MASAC-based high-level consensus planner. The planner generates smooth paths and achieves consensus among three agents under the 2-Neighbor communication topology.

\subsection{Low-Level and High-Level Controller Integration}
\begin{figure}[H]
\centering
\includegraphics[width=0.4\textwidth]{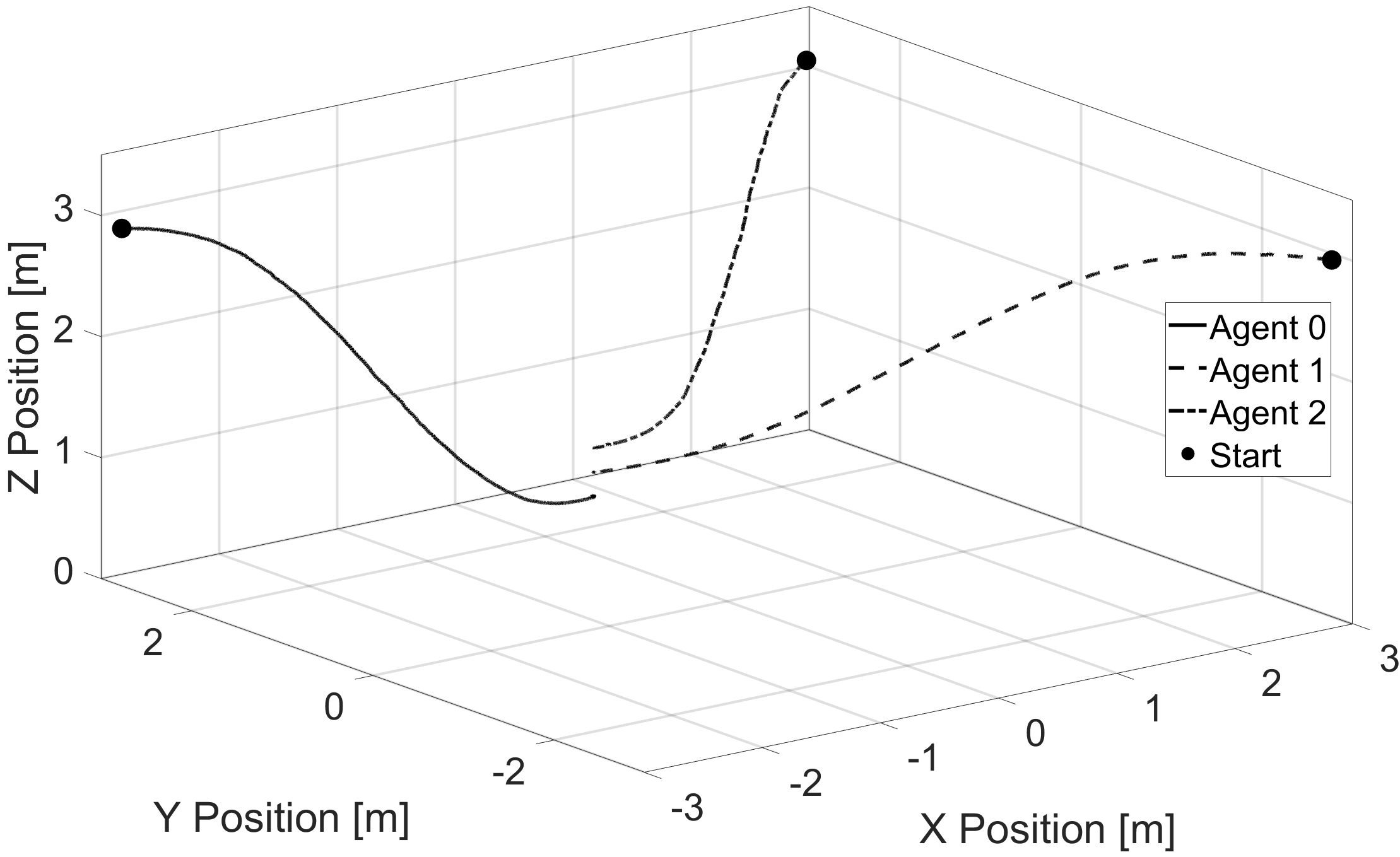}
\caption{Closed-loop response of three quadrotors converging to a common position under the proposed hierarchical architecture, combining a thrust-vector low-level tracking controller with the MASAC high-level consensus planner}
\label{fig:sac_3d}
\end{figure}

Fig.~\ref{fig:sac_3d} shows the closed-loop response of the integrated architecture in Fig.~\ref{fig:rl_stack}, where the MASAC consensus planner generates reference trajectories tracked by a thrust-vector low-level controller for three quadrotors. The planner produces smooth commands, and the tracking layer remains stable, indicating effective planner-controller compatibility and coordinated vehicle-level behavior.

\subsection{Distributed MARL vs. Centralized MARL}
Both the distributed MARL controller and the centralized MARL controller were trained on 3 agents. Both controllers were tested in 10 different runs. The distributed controller achieved a mean time-to-consensus of $3.7~\mathrm{s}$, while the centralized baseline converged in $3.6~\mathrm{s}$. The mean terminal consensus error (RMSE to the swarm mean position) was $0.001~\mathrm{m}$ for the distributed controller and $0.007~\mathrm{m}$ for the centralized controller, corresponding to a negligible absolute difference of $6~\mathrm{mm}$. As shown, the distributed MARL approach achieved similar results to the centralized MARL. 

However, the centralized MARL operates with an action space of a complexity of $O(N)$, and an observation space of a complexity of $O(N^2)$. Increasing the number of agents causes a large blowup in the policy and critic networks needed to be trained, increasing memory requirements and making accurate $Q$-value estimation more difficult as the joint observation space grows. Table~\ref{tab:centralized} shows the steady growth of these dimensions even for moderate $N$. On the contrary, the proposed distributed approach keeps the observation and action complexity as $O(1)$, yielding constant per-agent computational cost and enabling scalable deployment without redesigning the policy architecture or retraining, as shown in Sec.~\ref {sec:scale}.

\begin{table}[H]
\centering
\caption{Scaling of centralized MARL joint observation and action dimensions with team size $N$. The centralized formulation requires an $N^2 + N$ dimensional observation and a $2N$ dimensional joint action, leading to rapidly increasing policy/critic sizes as $N$ grows.}
\label{tab:centralized}
\begin{tabular}{ccc}
\toprule
\textbf{$N$} & \textbf{Observation} $\big(N^2+N\big)$ & \textbf{Action} $\big(2N\big)$ \\
\midrule
3   & 12  & 6 \\
5  & 30  & 10 \\
7  & 56  & 14 \\
10  & 110 & 20 \\
50  & 2,550 & 100 \\
100  & 10,100 & 200 \\
\bottomrule
\end{tabular}
\end{table}

\subsection{Consensus Particle Planner Scalability Test with 2-Neighbor Communication}
\label{sec:scale}
To assess scalability and zero-shot generalization under sparse communication, the original set of the three learned policies was evaluated and deployed to larger swarms without any retraining, fine-tuning, or adaptation. For each team size $N$, the three policies are reused and assigned to agents in a round-robin manner while maintaining a fixed 2-Neighbor communication topology. A total of 10 rollouts are executed per $N$, and the following metrics are reported:
\begin{enumerate}
    \item time-to-consensus (TTC), in minutes, measured when the swarm steady-state spread is constant within an $\epsilon = 1e-3$ for 5 seconds
    \item terminal consensus error (TCE) measured as RMSE to the group mean position in meters
\end{enumerate}

\begin{figure}[H]
\centering
\includegraphics[width=0.5\textwidth]{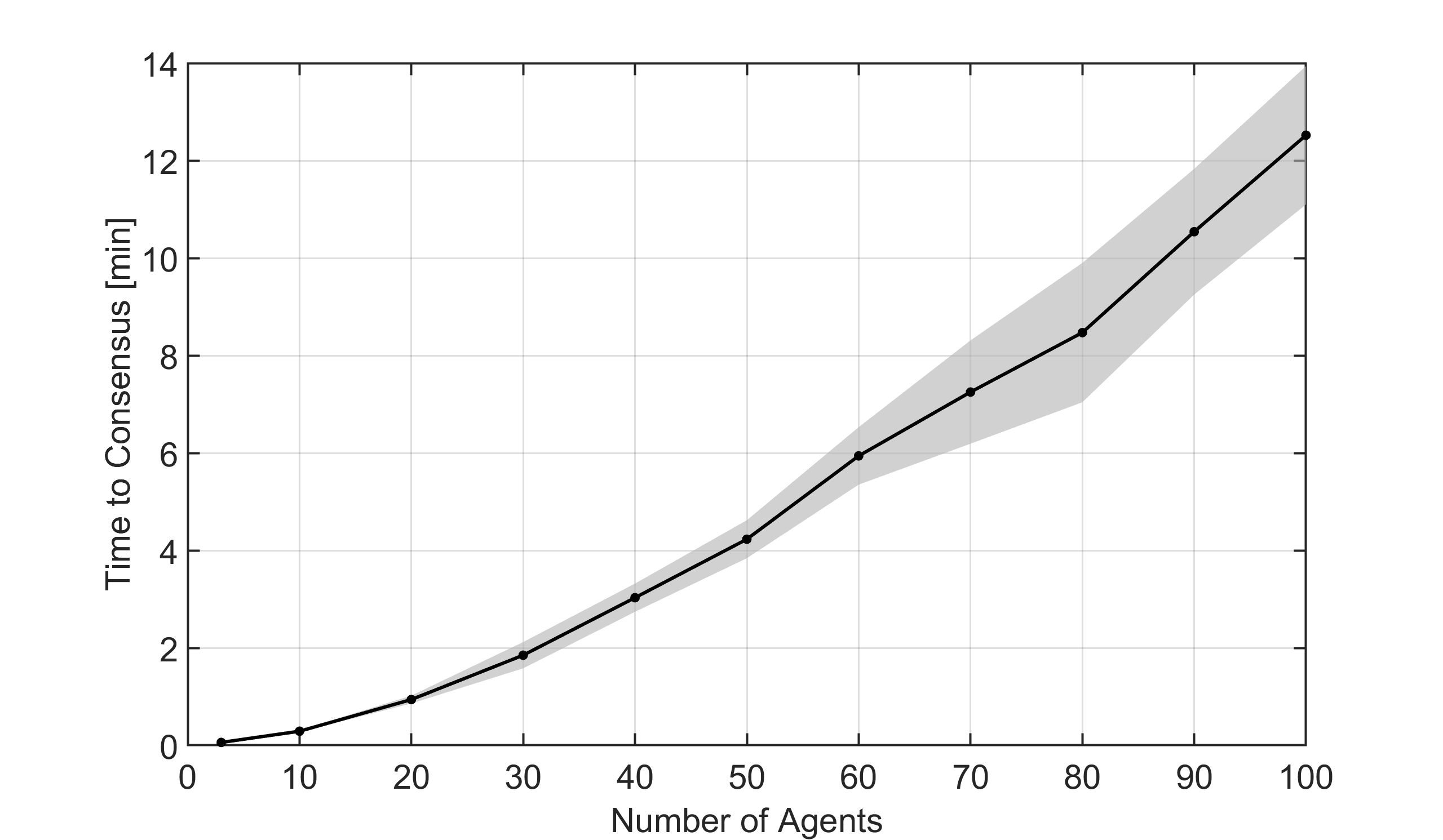}
\caption{Mean time to consensus (TTC) as a function of the number of agents under a 2-Neighbor communication topology. The shaded region indicates the standard deviation over multiple rollouts}
\label{fig:TTC}
\end{figure}

Fig.~\ref{fig:TTC} shows the mean time to consensus with the standard deviation across each number of agents. As shown, the mean time of consensus increases as the number of agents increases, with the standard deviation increasing as well. This is expected as due to the low fixed communication degree as $N$ grows the time of information propagation grows.

Results up to $N = 250$ are reported as mean $\pm$ standard deviation over runs and summarized in Table~\ref{tab:consensus_scalability_2edge}. Notably, the controller scales from $N=3$ up to $N=250$ agents with no additional training. Overall, the results demonstrate strong zero-shot scalability of the learned consensus controller under a constant-degree 2-Neighbor communication topology.

As expected, the time to consensus increases with $N$ due to slower information propagation in sparser graphs. The TCE remains low and well-bounded in agents up to $N=100$ but increases slightly with $N=250$. This is due to the fact that with a fixed 2-Neighbor communication local neighborhoods may reach a cluster that is tight locally but not globally. Despite this degradation in precision, the results demonstrate that the controller maintains stable convergence behavior. Importantly, the learned local consensus rule generalizes to significantly larger swarms while preserving fully distributed execution and a constant per-agent communication cost.
\begin{table}[H]
\centering
\caption{Scalability of the consensus controller under 2-Neighbor communication using a fixed checkpoint trained on $N=3$ agents. The same three learned policies are deployed to larger teams without retraining or fine-tuning (round-robin assignment). Reported values are the mean $\pm$ standard deviation over 10 rollouts per team size.}
\label{tab:consensus_scalability_2edge}
\begin{tabular}{ccc}
\toprule
\textbf{$N$} & \textbf{TTC [min]} & \textbf{TCE [m]} \\
\midrule
3   & 0.06 $\pm$ 0.00  & 0.0013 $\pm$ 0.0000 \\
20  & 0.94 $\pm$ 0.08  & 0.0121 $\pm$ 0.0051 \\
50  & 4.23 $\pm$ 0.39 & 0.0530 $\pm$ 0.0235 \\
80  & 8.47 $\pm$ 1.43 & 0.1013 $\pm$ 0.0304 \\
100 & 12.52 $\pm$ 1.42 & 0.1603 $\pm$ 0.0547 \\
250 & 32.17 $\pm$ 10.44 & 0.6440 $\pm$ 0.1486 \\
\bottomrule
\end{tabular}
\end{table}

\section{Conclusion}
This paper introduced a Network Distributed Multi-Agent Reinforcement Learning (MARL) formulation for UAV consensus control that integrates the communication graph into the decision process. By factorizing transitions and rewards over local neighborhoods, the proposed framework enabled distributed execution with constant per-agent computation and communication, matching the practical constraints of large swarms. Experimental results show that a MASAC-based high-level consensus planner produced smooth trajectories, integrated stably with a low-level thrust-vector tracking controller, and, most importantly, exhibited strong zero-shot scalability. The same three policies trained only on a three-agent system generalizes to swarms up to 250 agents without retraining, maintaining low error up to $N = 250$ under sparse 2-Neighbor communication. These findings suggest that ND-MARL provides a principled and scalable foundation for distributed, communication-aware quadcopter consensus control.

\bibliographystyle{ieeetr}
\bibliography{references}
\end{document}